\documentclass[10pt,twocolumn,letterpaper]{article}

\usepackage{cvpr}
\usepackage{times}
\usepackage{epsfig}
\usepackage{graphicx, subfigure}
\usepackage{amsmath}
\usepackage{amssymb}
\usepackage[numbers]{natbib}
\usepackage{enumitem}
\usepackage{enumerate}
\usepackage{comment} 
\usepackage[noblocks]{authblk}
\usepackage[breaklinks=true,bookmarks=false]{hyperref}

\cvprfinalcopy % *** Uncomment this line for the final submission

\def\cvprPaperID{3064} % *** Enter the CVPR Paper ID here

\begin{document}

%%%%%%%%% TITLE
\title{Multi-Object Tracking with Multiple Cues and Switcher-Aware Classification}

\renewcommand\Authsep{\space\space\space\space}
\renewcommand\Authands{\space\space\space\space}
\author[1]{Weitao Feng}
\author[1,2]{Zhihao Hu}
\author[1]{Wei Wu}
\author[1]{Junjie Yan}
\author[3]{Wanli Ouyang}
\setlength{\affilsep}{1em}
\affil[1]{Sensetime Group Limited}
\affil[2]{Beihang University}
\affil[3]{The University of Sydney}
\affil[ ]{
\texttt{\{fengweitao, huzhihao, wuwei, yanjunjie\}@sensetime.com}
\authorcr
\texttt{wanli.ouyang@sydney.edu.au}
}
%\author{Weitao Feng\\
%SenseTime Group Limited\\
%% fengweitao@sensetime.com\\
%{\tt\small fengweitao@sensetime.com}
%% For a paper whose authors are all at the same institution,
%% omit the following lines up until the closing ``}''.
%% Additional authors and addresses can be added with ``\and'',
%% just like the second author.
%% To save space, use either the email address or home page, not both
%\and
%Zhihao Hu\\
%SenseTime Group Limited\\
%Beihang University\\
%{\tt\small huzhihao@sensetime.com}
%\and
%Wei Wu\\
%SenseTime Group Limited\\
%%Address line\\
%{\tt\small wuwei@sensetime.com}
%\and
%Junjie Yan\\
%SenseTime Group Limited\\
%%Address line\\
%{\tt\small yanjunjie@sensetime.com}
%\and
%Wanli Ouyang\\
%The University of Sydney\\
%%Address line\\
%{\tt\small wanli.ouyang@sydney.edu.au}
%}

\maketitle
%\thispagestyle{empty}

%%%%%%%%% ABSTRACT
\begin{abstract}
    
    In this paper, we propose a unified Multi-Object Tracking (MOT) framework learning to make full use of long term and short term cues for handling complex cases in MOT scenes.  Besides, for better association, we propose switcher-aware classification (SAC), which takes the potential identity-switch causer (switcher) into consideration. Specifically, the proposed framework includes a Single Object Tracking (SOT) sub-net to capture short term cues, a re-identification (ReID) sub-net to extract long term cues and a switcher-aware classifier to make matching decisions using extracted features from the main target and the switcher. Short term cues help to find false negatives, while long term cues avoid  critical mistakes when occlusion happens, and the SAC learns to combine multiple cues in an effective way and improves robustness. The method is evaluated on the challenging MOT benchmarks and achieves the state-of-the-art results. 

\end{abstract}
%%%%%%%%% BODY TEXT
\section{Introduction}
Multi-Object-Tracking (MOT) is important in video analysis systems, such as video survelliance and self-driving car. It aims to maintain trajectories of all targets from categories of interest.  The most recent methods in MOT follow the tracking-by-detection paradigm, which takes the frame-wise detections as the input and associates detections as the final trajectories. However, the detections are not always accurate enough, which could substantially influence the tracking. Besides, the occlusion and abnormal motion are another two problems in MOT.

\begin{figure}[t]
\begin{center}
  \subfigure[]{
    \label{fig:subfig:b} %% label for second subfigure
    \includegraphics[width=1.2in]{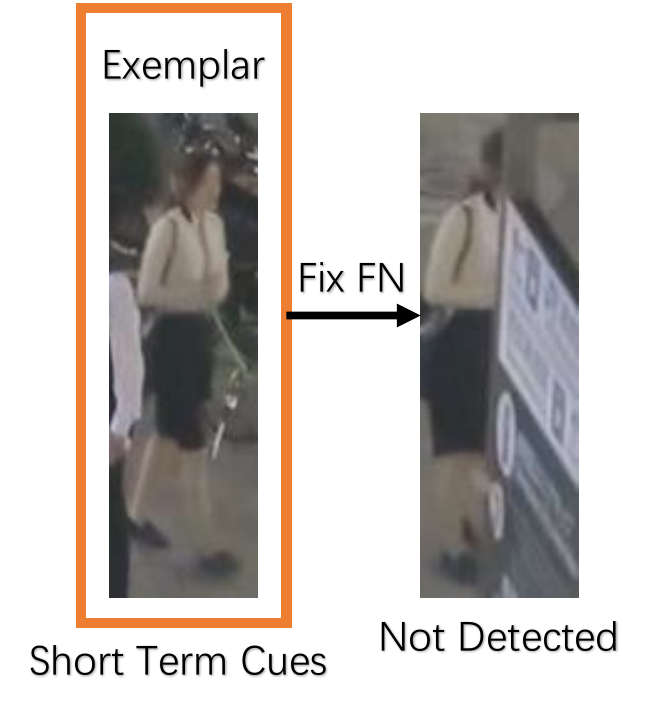}}
   \subfigure[]{
    \label{fig:subfig:a} %% label for first subfigure
    \includegraphics[width=1.95in]{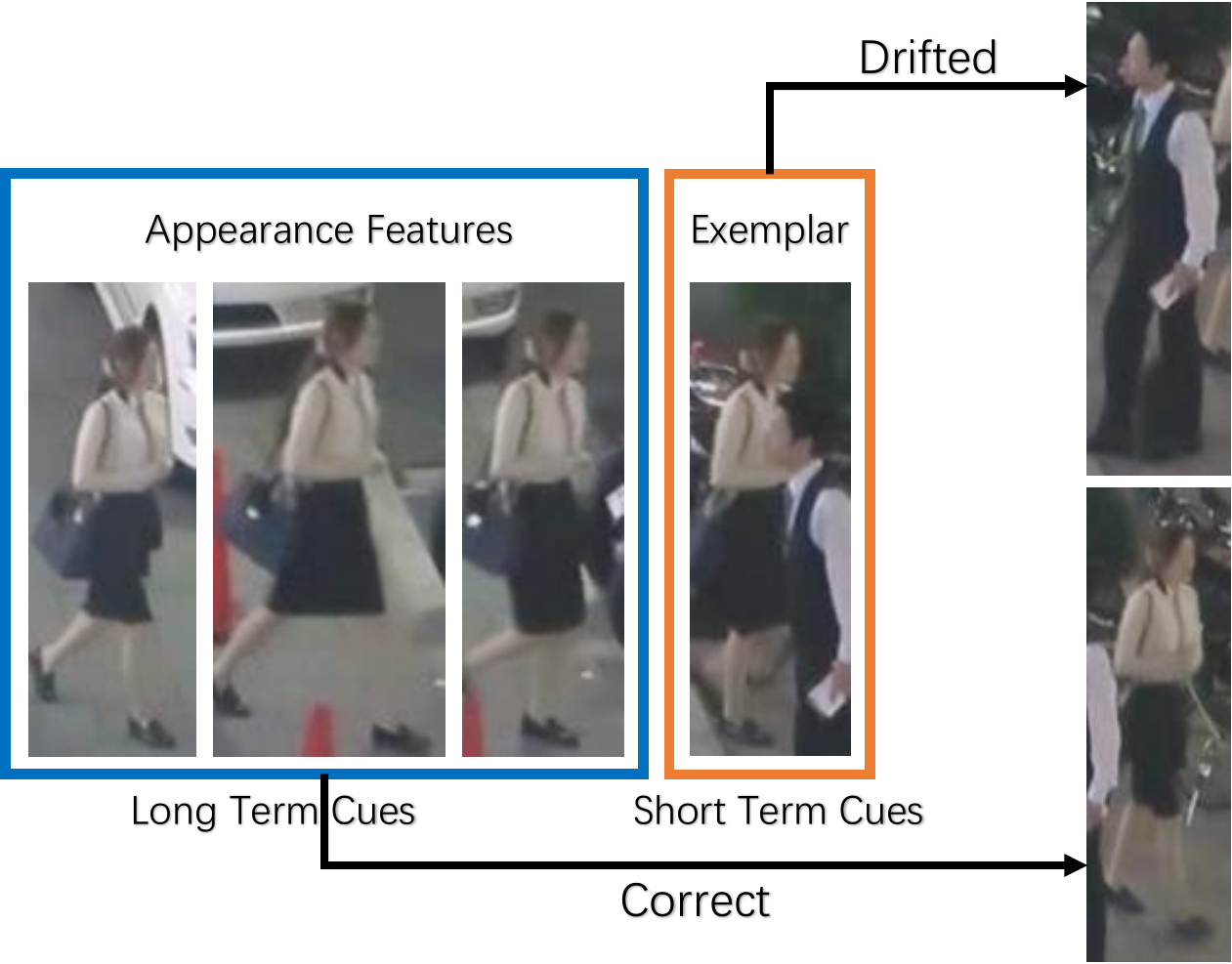}}
  \subfigure[]{
    \label{fig:subfig:c} %% label for second subfigure
    \includegraphics[width=2.8in]{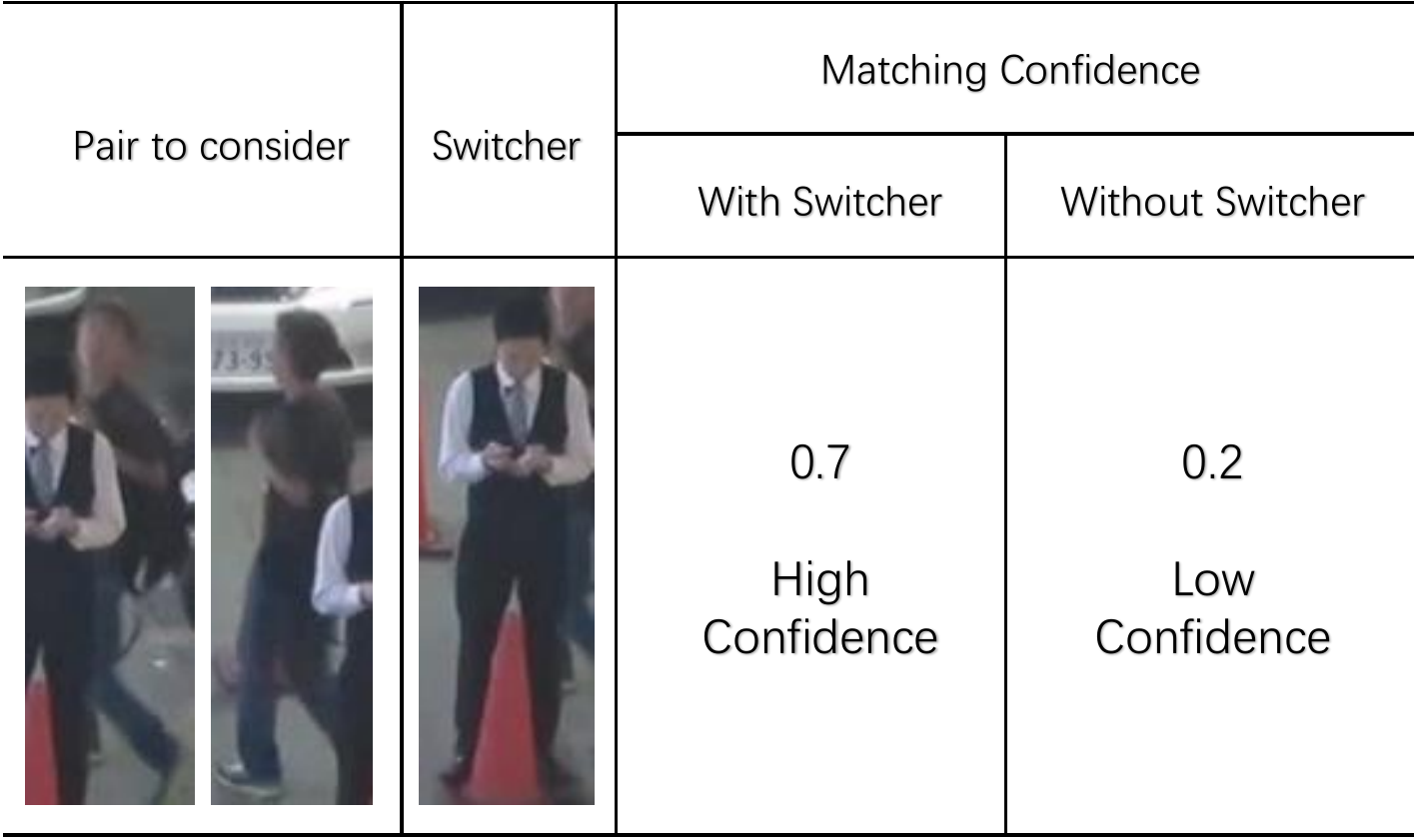}}
\end{center}
   \caption{ (a) False negative (FN): detector may not detect the occluded target, while SOT tracker can find this target to complement the detector. (b) Occlusion: when occlusion happens, the updated appearance is similar to the wrong target, and SOT is easy to drift. In contrast, the overall appearance of the tracklet is still stable and more reliable for association. (c) Switcher helps matching: without switcher, the matching confidence for the correct pair is low, while the matching score is higher when the classifier is aware of the occlusion situation and appearance information of the switcher. }
%\label{fig:long}
\label{fig:fig1}
\end{figure}
We define two different cues in MOT. The short term cues mean updated cues between neighbouring frames, which include current target position, appearance and motion. The long term cues stand for tracklet-long cues containing appearance features of the object within the tracklet.  The recent single object tracking (SOT) approaches with high performance can be used in MOT for capturing short term cues, which are helpful when handling inaccurate detection results and abnormal motion. As shown in Figure~\ref{fig:subfig:b}, SOT trackers are effective to reduce false negatives (FN). Though short-term cues are helpful in many cases, most short term cues become unreliable when occlusions happen because the inclusion of occluded region makes the SOT tracker drift. Then long term cues of tracklet appearance can help to avoid the drift in SOT caused by occlusion.  For the example shown in Figure~\ref{fig:subfig:a}, long term cues are still stable when occlusion happens. 

Previous works did not make the fully use of the two cues. Many works like \cite{zhu2018online} that include SOT tracker in MOT did not consider combining SOT results in data association, while some works like \cite{sadeghian2017tracking} did not use SOT trackers to handle short term cues. Other rule based combination of long term and short term cues like \cite{yoon2018online} cannot learn from data with different situations and may over-fit to some specific cases. It also is a question for learning effective combination of short term and long term cues. In previous experiments (see Sec.~\ref{sec:abla}), we have observed that it is hard to combine all cues in one network. That is, SOT tracker for short term cues cannot distinguish similar objects, and the network for long term cues cannot predict the precise position of target. Based on this motivation, we propose a unified MOT framework to generate short term and long term cues, as well as adaptively choose them for data association. 

Another motivation of this paper is to use local interaction information to solve identity-switches.
%%use ident ity-switch causer (switcher) as the extra information for data association.
%%Ignorance to local interaction is one of the reasons why we can hardly notice critical identity-switches.
We have found that the potential identity-switch causer (switcher) is critical for correct matching.
For example, Figure~\ref{fig:subfig:c} shows how the switcher helps matching. Driven by this motivation, we use a switcher-aware classifier, which is implemented using boosting decision trees, to encode potential switcher information and improve the tracking robustness.

This paper proposes MOT approach with multiple cues and switcher-aware classification.
The state-of-the-art method of SOT is used for capturing short term cues and a re-identification (ReID) method is applied for extracting long term cues. 
During data association, the switcher-aware classifier gathers all long term and short term cues and takes potential switcher into consideration, then generates scores building a bipartite graph for matching. 

%% \begin{figure}[t]
%% \begin{center}
%% %\fbox{\rule{0pt}{2in} \rule{0.9\linewidth}{0pt}}
%%    \includegraphics[width=0.8\linewidth]{fig-a.png}
%% \end{center}
%%    \caption{(a) Occlusion: short-term cues become unreliable, while long-term cues can help fix mistake. (b) False negative: short-term cues can %% reduce missing detections.  }
%% %\label{fig:long}
%% \label{fig:fig1-a}
%% \end{figure}
%% 

The main contributions of our work are listed as follows:
% \begin{itemize}[noitemsep,nolistsep]

1. An effective MOT framework learning to capture long term and short term cues and making adaptive decisions for robust tracking. %The framework is data-driven, and is efficient for data association even when handling large amount of targets. 

2. A switcher-aware classification (SAC) in data association for improving the robustness of MOT to identity switch. We also introduce a simple but effective approach to search for potential switcher.

Extensive experimental results on both MOT16 and MOT17 benchmarks\cite{MOT16} clearly show the effectiveness of the proposed framework.

% \end{itemize}

%\textcolor{red}{Can we say “Experimental results show that the proposed framework improves ?? by ?? on MOT16 and ?? on MOT ?? when compared with existing best results"?}

%------------------------------------------------------------------------
\section{Related Work}

\begin{figure*}
\begin{center}
%\fbox{\rule{0pt}{2in} \rule{.9\linewidth}{0pt}}
\includegraphics[width=6.0in]{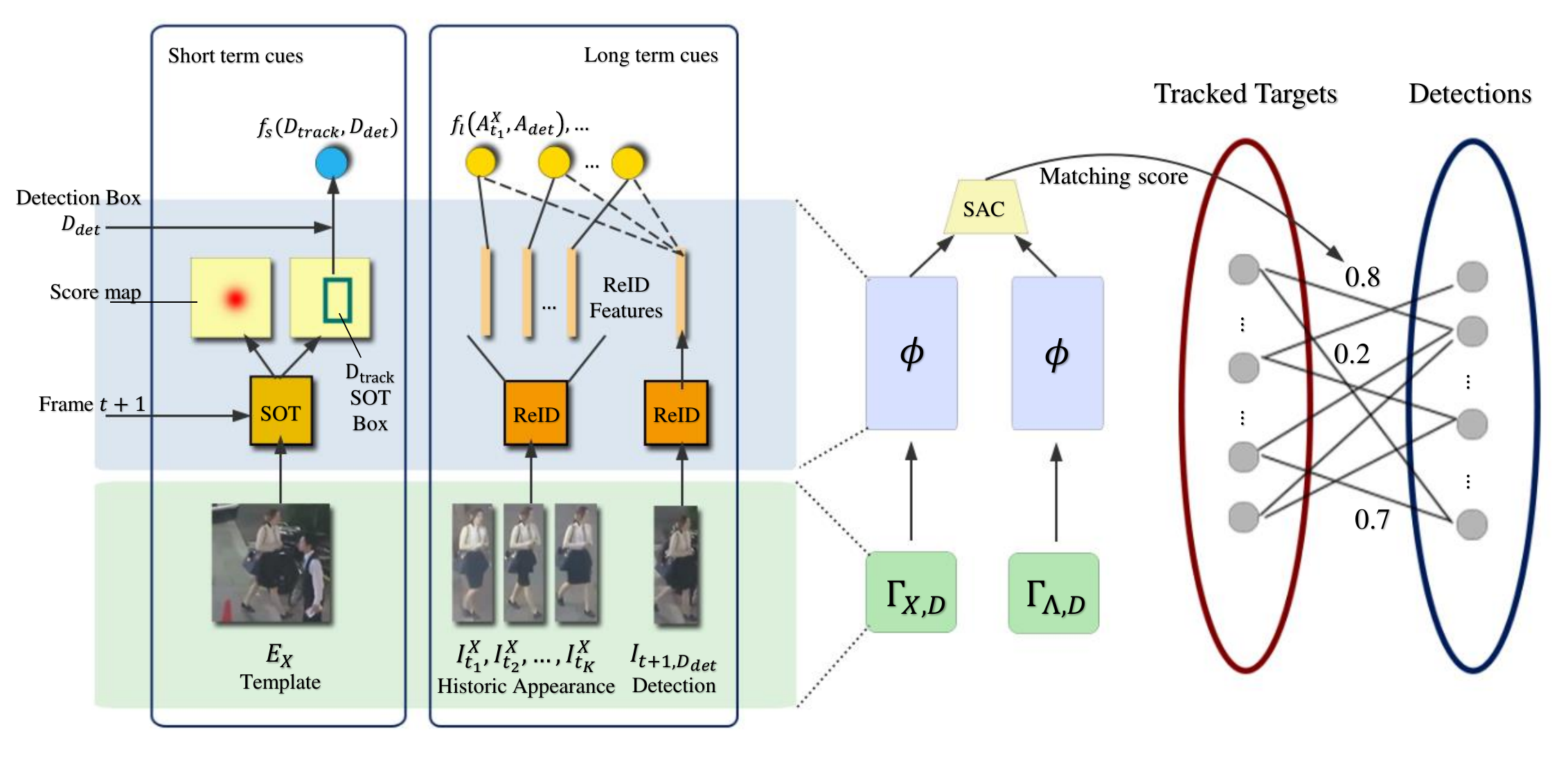}
\end{center}
   \caption{Overall design of the proposed MOT framework. Short term cues are captured by SOT sub-net and long term cues are captured by ReID sub-net. The switcher aware classifer (SAC) predicts whether the detection result matches the target using the short term and long term cues of the target, the detection result, and the switcher.}
\label{fig:arch}
\end{figure*}

%-------------------------------------------------------------------------
\subsection{MOT Using SOT Tracker}
Some previous works \cite{zhu2018online,chu2017online,xiang2015learning,yan2012track} have tried to apply SOT trackers into MOT task. 
Chu \etal \cite{chu2017online} uses CNN-based single object tracker and handles drift through a spatial-temporal attention mechanism, it regards all detections as SOT proposals. Xiang \etal \cite{xiang2015learning} utilizes MDP method to track targets in tracked state with optical flow.
Most works have never benefited from the progress of visual object tracking (VOT) task.
In recent years significant progress has been made in the single object tracking field. Trackers like GOTURN\cite{held2016learning}, Siamese-FC\cite{bertinetto2016fully}, ECO\cite{danelljan2017eco}, Siamese-RPN\cite{li2018high} have highly improve the tracking accuracy. 
Method proposed by \cite{zhu2018online} directly applies the ECO-HC\cite{danelljan2017eco} tracker from visual object tracking task with a cost-sensitive loss and designed a spatial-temporal network for data association when SOT tracker is considered losing the target.  However, an online-updating SOT tracker is slow in speed and costs a lot of memory. While offline training siamese SOT trackers like Siamese-RPN\cite{li2018high} reach the state-of-the-art accuracy at a high speed of more than 80 frames(targets) per second.
More importantly, most methods have not combined cues generated from SOT tracker with other cues. They separate the SOT tracker and the data association process.
%\textcolor{blue}{(DELETE?) Different from these works, we apply advanced offline-trained SOT tracker as a component, and include the cues from SOT when doing data association. In occlusion handling, we exploit long term ReID features instead of complicated spatial-temporal attention mechanism, which may be more simple but effective.} 
%\textcolor{red}{(CHECK IF OK) 
Different from these works, we use the information from SOT together with long term cues from the trajectory for learning to associate detection/tracking results. Our usage of long term cues helps to solve the problems of drift in SOT, which cannot be solved  effectively in existing SOT for MOT approaches.
%}

%-------------------------------------------------------------------------
\subsection{Data Association}
Data association is an important procedure of all tracking-by-detection-based MOT methods. 
\cite{zhang2008global,tang2015subgraph,tang2016multi,tang2017multiple,pirsiavash2011globally} formulate the data association process as various optimization problems. Most of them are variants of graph segmentation problem and they need batch processing. Most online processing methods use Hungarian Algorithm\cite{Munkres1957Algorithms} or minimum-cost-network-flow to solve a bipartite graph matching problem and they are effective.

Some works like \cite{son2017multi, sadeghian2017tracking, milan2017online, huang2008robust, bae2014robust, bae2018confidence} emphasize to improve the features used in data association.
\cite{sadeghian2017tracking,milan2017online,kim2018multi} exploit RNNs in the MOT task. Sadeghian \etal \cite{sadeghian2017tracking} combines appearance, motion and interaction cues into a unified RNN network. Milan \etal \cite{milan2017online} focuses on the utilization of positions and motions of the targets. Son \etal \cite{son2017multi} develops a new training method with ranking loss and regression loss to obtain higher accuracy. 

However, the disadvantages of these methods are well-known. First, they did not use SOT tracker to deal with inaccurate detection results, especially false negatives. Second, although they try a lot of discriminative features, they seldom take the local position and interaction information into training or inference phase. 
Sadeghian \etal \cite{sadeghian2017tracking} use spatial information from neighboring identities. But they do not use SOT tracker or the appearance information from switcher.
Some works like \cite{long2018real, yoon2018online} ensemble appearance features with position and motion features. Their designs in using these information sources are based on heuristic rules, but not based on principle learning.
Most of these works just regard data association as a pairwise matching problem between single tracklet and the detection. It is obvious that we will lose the valuable local interaction information which may indicate some critical errors. Though these mistakes will not cause huge drawback on target recall or precision, e.g., not a drawback in primary metric MOTA\cite{bernardin2008evaluating}, somehow they influence the robustness of a tracking system and are significant to its application. In this work, we introduce the SAC for robust tracking which takes the most possible switcher into training and inference. And we combine the short term cues and long term cues in a balanced and data-driven way. 

%\textcolor{red}{discuss: (this seems to be the advantage, can be put in introduction or methods)}
%

%------------------------------------------------------------------------
\section{The Proposed Framework}
%-------------------------------------------------------------------------
% In this section, we first introduce the overall design and the two sub-nets to obtain short term and long term cues. Then we introduce the method we generate the input of the classifier. As for the combination of short term and long term cues, we have found that the classifier input data size for each part should be balanced. We then replace the raw ReID features with similarity scores, and it is still effective to combine using such small features. At last, we introduce the training phase, pre-process and post-process procedures.

 \begin{bf}
 Problem Formulation. 
 \end{bf} The trajectory of one tracked target can be denoted by $ X = \{X_t\}$, where $ X_t = [X_t^x, X_t^y, X_t^w, X_t^h]$, $t$ is the frame index, $X_t^x, X_t^y$ is the top-left coordinate of the bounding box, and $X_t^w, X_t^h$ is the width and height of the bounding box. $q_X$ is the overall tracking quality for target $X$ and $I_t^X$ is the appearance of target $X$ at frame $t$. 
 
\subsection{Overall Design}
\label{sec:ov}
Figure~\ref{fig:arch} shows the overall design of the proposed MOT framework.
The framework uses the following steps for online mode:
%\textcolor{red}{Introduce how to get tracked target and its tracklet. Detection -> SOT plus overlap for finding candidate pair -> Data association for matching the previous trajectory with the detection results at the current frame.}

\begin{itemize}[leftmargin=12pt,noitemsep,nolistsep]
    \item Step 1. Initially, the set $\mathcal{S}$ of tracked targets is empty and $t=1$. 
    \item Step 2. At frame $t$, for a target $X$, the template $E_X$ of the target is sought in the next frame $I_{t+1}$ by using the SOT sub-net. The SOT sub-net outputs the most possible location $D_{track}$ for the template in $I_{t+1}$.
    \item Step 3. %The set of possible detection results in $I_{t+1}$ that could match the target at frame $t$ is constructed. 
    For a detection result $D_{det}$ in $I_{t+1}$, its corresponding detection image region $I_{t+1, D_{det}}$ and the historical appearances of the target $\{I_{t_{i}}^X\}, i=1,2,...,K$ will be used by the ReID sub-net to extract the long-term ReID features.
    \item Step 4. The location of the target $D_{track}$ found by SOT in Step 2, the location $D_{det}$ found by the detector, the ReID features obtained in step 3 will be combined into the matching feature of the target.
    \item Step 5. Find the potential switcher of the target $\Lambda$, i.e., the most possible identity switch causer, and extract its SOT and ReID features.
    \item Step 6. With the aid of the matching features of the switcher, the matching features of the target are used by the switcher-aware classifier (SAC) to generate the matching score on whether the detection result should match the target or not. This step is repeated for each detection result in the frame $I_{t+1}$ in order to obtain their matching scores to the tracked target. %The SAC is denoted by $\Omega$.
    \item Step 7. Build a bipartite graph of tracked targets and the detection results using the matching scores between the tracked targets and the detection results found in Step 6. Find the matching results of the graph using minimum-cost-network-flow.
    \item Step 8. For the matched targets, update the position and template using the information of the matched detection. For targets not matched, update tracklet position using SOT results, and drop targets which are considered unreliable or lost. For isolated detection results, if their confidence scores satisfy the condition of new target, they will be added to the set of tracked targets. 
    \item Step 9. Repeat steps from 2 to 8 for the next frame by setting $t=t+1$, until no more frames arrive.
\end{itemize}

% For $i$-th detection $D_{t+1}^i$ in frame $t+1$ and target tracklet $X$ along with potential switcher of $X$ at frame $t$ denoted by $\Lambda(X, t)$, the framework input can be denoted by $A_X^D = [X_t, X_{t-\delta1} X_{t-\delta2}, \Lambda(X, t)_t, \Lambda(X, t)_{t-\delta1}, \Lambda(X, t)_{t-\delta2}, D_{t+1}^i]$. The predicted matching possibility $P_X^D = \phi(A_X^D;W)$. $\phi(\cdot;W)$ is the framework architecture.
\subsection{Using SOT Tracker for Short Term Cues}

\begin{figure}
\centering
%\fbox{\rule{0pt}{2in} \rule{.9\linewidth}{0pt}}
  \subfigure[]{
    \label{fig:side:a} %% label for first subfigure
    \includegraphics[width=3.2in]{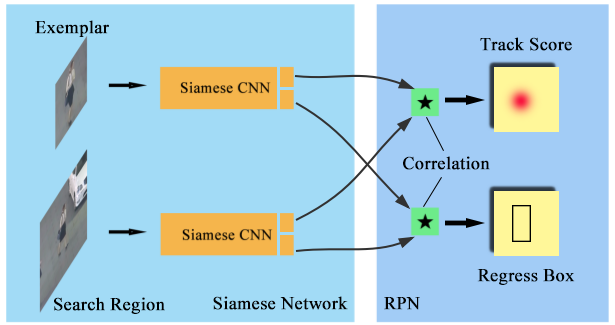}}
%   \subfigure[]{
%     \label{fig:side:b} %% label for first subfigure
%     \includegraphics[width=3.35in]{rearch2.png}}
  \caption{Siamese-RPN architecture for SOT.}
\label{fig:sot}
\end{figure}

\begin{bf}
Baseline tracker. 
\end{bf}
We use Siamese-RPN tracker \cite{li2018high} in our framework for extracting short term cues. Following the original schema, the template $E_X$ at the current frame, also called the exemplar, is resized to $127\times 127$. To search the target at the next frame, search region $R$ is cropped from frame $I_{t+1}$ according to the position of $X$.  The search region is then resized to $255\times 255$. Specifically, the picture scale of the search region is the same as the exemplar. As shown in Figure~\ref{fig:sot}, the search region and the exemplar are passed through the shared-weight siamese CNN. The CNN features of the exemplar are then used by two branches, each consisting of two convolution layers, similarly for the search region. One branch for obtaining the score maps and the other for bounding box regression. The correct location of the exemplar in the search region is supposed to have largest score in the score map. The bounding box regression at different locations should point to this correct location.
% \textcolor{red}{The exemplar features from the CNN are then up-channelled to two branches of $2k$ and $4k$ times channel size through two extra convolution layers. 
% The features of search region are passed though\textcolor{blue}{through} another two extra convolution layers to generate two search region branch\textcolor{blue}{es} but keep channel size unchanged.    
% Here $k$ is the number of anchors. 
% After that, the features from the exemplar and one of the search region branches are combined by correlation layer and used for binary classification. The $4k$ branch and the rest one of search region branches are for bounding box regression.  }
%\textcolor{red}{not clear: Features extracted from the shared-weight CNN are fed to another 4 convolution layers to create 4 branches features, i.e., two for exemplar and two for search region. Through two correlation operations, the 4 branches finally form two output, one for binary classification, and the other for bounding-box regression.}

\begin{bf}
Short term features generation.
\end{bf}
The SOT sub-net outputs a SOT score and the predicted bounding box, called SOT box. The detection bounding box to be matched is called detection box. We denote the SOT score as $p$, the SOT box as $D_{track}$, the detection box as $D_{det}$, then short term feature $f_s$ is calculated as following:
\begin{equation}
\begin{split}
    f_s(D_{track}, D_{det}) = IoU(D_{track}, D_{det})
\end{split}
    \label{eq:short}
\end{equation}

\begin{bf}
Distractor-aware SOT tracker for MOT tracking.
\end{bf}
In order to maximize the effect of Siamese-RPN. We modify the anchors to fit the target scales of pedestrian. Besides, we refine the network using pedestrian data. 
Another problem of the SOT tracker is that it is hard to tell when to stop tracking if the target is lost. When the tracker drifts to background distractor, the tracker may not be able to stop tracking the distractor.
To make the tracker score distractor-aware, we design a tracking score refine strategy. We refine the tracker score using the matching results found at the step 7 introduced in Sec.~\ref{sec:ov}.
For target $X$, the refined overall tracking quality $q_X$, is as following:
\begin{equation}
\begin{split}
    q_{X, t+1} = \left\{
             \begin{array}{rcl}
             \frac{q_{X, t} + IoU(D_{track}, D_{det})\cdot p}{2},  \textrm{if\  matched,}\\
              q_{X, t} \cdot decay \cdot p^k,  \textrm{otherwise,}
             \end{array}
    \right.
\end{split}
    \label{eq:trackingq}
\end{equation}
where $decay$ and $k$ are hyper-parameters used for dealing with inconsistent targets, $D_{det}$ is the detection box. In this way, we can drop unreliable targets if the tracking quality $q_X$ is below a threshold $\zeta_{t}$.
%-------------------------------------------------------------------------
\subsection{Using ReID Network for Long Term Cues}
We use a modified version of GoogLeNet Inception-v4 as the backbone CNN of the ReID sub-net. 
The ReID feature is extracted from the last FC layer before classification. Table~\ref{tab:incepti} and Table~\ref{tab:incblock} show details of the backbone CNN.
 
\begin{table}
\footnotesize
\begin{center}
\begin{tabular}{|l|c|c|c|}
\hline
Type & Kernel/Stride & Output Size & Padding \\
\hline\hline
input & - / - & $3\times224\times224$ & - \\
convolution & $7\times7$ / 2 & $29\times112\times112$ & 3 \\
pool & $3\times3$ / 2 & $29\times56\times56$ & - \\
convolution & $1\times1$ / 1 & $27\times56\times56$ & - \\
convolution & $3\times3$ / 1 & $142\times56\times56$ & 1 \\
pool & $3\times3$ / 2 & $142\times28\times28$ & - \\
inception-A & - / - & $379\times28\times28$ & - \\
inception-A & - / - & $679\times28\times28$ & - \\
pool & $3\times3$ / 2 & $679\times14\times14$ & - \\
inception-A & - / - & $1037\times18\times18$ & - \\
inception-A & - / - & $1002\times18\times18$ & - \\
inception-A & - / - & $938\times18\times18$ & - \\
inception-A & - / - & $861\times18\times18$ & - \\
pool & $14\times14$ / 1 & $861\times1\times1$ & - \\
fc & - / - & 256 & - \\
\hline
\end{tabular}
\end{center}
\caption{The modified Inception-v4 architecture.}
\label{tab:incepti}
\end{table}
\begin{table}
\footnotesize
\begin{center}
\begin{tabular}{|l|c|c|c|c|}
\hline
Step & Branch A & Branch B & Branch C & Branch D \\
\hline\hline
1 & conv $1\times1$ & conv $3\times3$ & conv $3\times3$ & conv $1\times1$ \\
2 & - & conv $1\times1$ & conv $3\times3$ & pool $3\times3$ \\
3 & - & - & conv $1\times1$ & - \\
\hline
\end{tabular}
\end{center}
\caption{One inception-A block: all convolution layers of $3\times3$ and the pooling layer use padding 1, the others have no padding, and the pooling layer stride 2.}
\label{tab:incblock}
\end{table}

\begin{bf}
Quality-aware long term tracklet history construction. 
\end{bf}
To select $K$ images from the tracklet history of the target, we design a quality-aware mechanism.
In order to get long term cues of good quality, we use a quality filter to select the best $K$ images in the past $K$ time periods to ensure quality and robustness. The indices of the $K$ frames selected as the tracklet history of the target are denoted by $\mathcal{H}=\{t_1 \ldots, t_K\}$. The $K$ frames we choose are defined below:
\begin{equation}
    t_i = \mathop{\arg\max}_{t - i\delta < \hat{t} \leq t - (i-1)\delta} \mathop{Q}(I_{\hat{t}}^X) , i = 1, 2, ..., K
    \label{eq:select}
\end{equation}
where $Q$ is a network which outputs the quality score. $Q$ is implemented using a Resnet-18 model. $I_{t}^X$ is the image region of target $X$ at frame $t$.   $\delta$ is hyper-parameter deciding the selecting interval.
For example, when $i=1$, $t_1$ is  chosen from the frame with maximum quality score among frames $t, t-1, ..., t-\delta+1$. When $i=2$, $t_2$ is chosen from $t-\delta, t-\delta-1, ..., t-2\delta+1$. Therefore, the $i$ in $t_i$ corresponds to different stride and search range.

\begin{bf}
Long-term feature generation. 
\end{bf}
After the $K$ images are selected, all these images and the detection result to be matched will be fed to the ReID sub-net and output their ReID features. Then we can obtain $K$ long term features for the target as follows:
\begin{equation}
\begin{split}
&\mathcal{F}^X_l = \{f_l(A_{t_i}^X A_{det})| i=1, \ldots, K\}, \\
&\textrm{where }    f_l(A_{t_i}^X, A_{det}) = \frac{\strut{A_{t_i}^X} ^\textrm{T} \cdot A_{det}}{\left | A_{t_i}^X \right | \left | A_{det} \right |},
\end{split}
    \label{eq:cosd}
\end{equation}
$A_{t_i}^X$ is the vector containing ReID features of the $i$th image selected from tracklet history of target $X$, $A_{det}$ is the ReID features of the detection result. 
To save computation, each image in the tracklet will have their features extracted by the ReID network only once. Their features are saved for further calculation.

%-------------------------------------------------------------------------
\subsection{Switcher-Aware Classifier}

\begin{bf}
Switcher retrieval.
\end{bf}
We have observed large amount of identity switches (IDS) and find that most IDSs occur when two targets meet each other with large overlap. It inspires us to mark the other target having the largest overlap with current one as the most possible potential switcher. Mathematically, for each tracklet $X$ in frame $t$, its position is denoted by $X_t$, and the potential switcher is obtained as follows
\begin{equation}
    \Lambda = \mathop{\arg\max}_{Y \in \mathcal{S}  \mathop{s.t.} Y \neq X} \mathop{IoU}(X_t, Y_t).
    \label{eq:swre}
\end{equation}
where $\mathcal{S}$ is the set of tracked targets.

\begin{bf}
Input features.
\end{bf}
Here we consider the two sub-nets as a feature extraction operator $\phi$, and denote the input of the two sub-nets for target $X$ and detection result $D$ as $\Gamma_{X,D}$, similarly for the switcher.
The input features of the classifier consist of two parts: the features of mainly considered target, denoted by $\phi(\Gamma_{X,D})$, and the features of the switcher, denoted by $\phi(\Gamma_{\Lambda,D}))$. $\phi$ is defined bellow. 
%\textcolor{red}{define $\Gamma$ when it first appears}

\begin{equation}
\begin{split}
\phi(\Gamma_{X,D})&= \{f_s(D_{track}, D_{det})\} \cup \mathcal{F}^X_l.
\end{split}
    \label{eq:input}
\end{equation}
The dimension of $\phi(X, t)$ is $K+1$, and similarly for the switcher. Then we obtain the input of the classifier by concatenating the two parts.
% With tracklet $X$ and corresponding switcher $\Lambda(x, t)$ as framework input, we extract their predicted IoU using SOT sub-net as short term feature, sample three checkpoint from tracklet history and pass them through ReID sub-net together with the detection then calculate the three cosine distance as long term features. Additionally, we take the tracklet length from tracklet status into consideration. We do the same on the switcher. Finally these features are concatenated and form the input feature of size 10. 
% \begin{equation}
% \begin{split}
%     Input = &[\\
%     &f_s(), cos_{t}, cos_{t-\delta1} , cos_{t-\delta2}, len, \\
%     &IoU^{sw}, cos_{t}^{sw}, cos_{t-\delta1}^{sw} , cos_{t-\delta2}^{sw}, len^{sw}\\
%     &]
% \end{split}
% \end{equation}

\begin{bf}
Classification.
\end{bf}
We exploit regularized Newton boosting decision tree with weighted quantile sketch, which is proposed by \cite{chen2016xgboost}, in the classification step. 
If the classification result $y$ is larger than threshold $\zeta_{m}$, then the corresponding edge with cost $1 - y$ will be added to graph.

\subsection{Training}
\subsubsection{Training data generation}

The SOT sub-net and the ReID sub-net are trained independently. For the SOT sub-net,  we generate some image pairs of targets according to the ground truth of the videos and the pairs are extended to include part of the background according to the training schema of Siamese-RPN. For better training, we only consider samples which have IoU to ground truth larger than 0.65 as positives and smaller than 0.45 as negatives. For ReID sub-net, each target is regarded as one class and we train the net to predict the class of the input target. The input of ReID sub-net is the target image region the label is its class number.

In order to generate training samples and corresponding annotations for the switcher-aware classifier, we should first generate the inputs of the two sub-nets.
At the beginning, we run a baseline MOT algorithm and generate all hypothetical tracklets of training videos. Then we associate them with the ground truth using an IoU threshold of $0.6$, i.e., only pairs with IoU larger than $0.6$ will be considered. The sum of IoU value is maximized by Hungarian Algorithm. 
For target $X$ at frame $t$, if the ground truth of $X_{t}, X_{t-\delta}, ..., X_{t-(K-1)\delta}$ belong to the same target or at most one of them is not associated with a ground truth, then we consider it as a valid tracklet. For each valid tracklet, we randomly sample a detection result in frame $t+1$, together with the potential switcher defined by Eq.~\ref{eq:swre}, the input of the MOT framework is done. According to Eq.~\ref{eq:short},~\ref{eq:cosd},~\ref{eq:input}, we can generate the inputs of the classifier.
According to the ground truth of the detection, we can obtain the label of the switcher-aware classifier. Besides, for a positive sample, we can exchange the switcher and the mainly considered one to generate another training negative sample for the switcher-aware classifier.

% \textcolor{red}{NOT CLEAR, $X_{t_1}$ NOT DEFINED: For every frame $t$ and every bounding-box $X_t$, if the ground truth of $X_{t_1}, X_{t_2}, ..., X_{t_K}$ are the same or at most one of them is none of the ground truth, then we consider it is a valid tracklet.} \textcolor{red}{RECONSIDER what are the input of the framwork: For each valid tracklet, we randomly sample a detection result in frame $t+1$, together with the potential switcher, the input of the framework is done.} According to the ground truth of the detection, we can obtain the label.

As for the quality filter, we generate target image regions for input with IoU to ground truth larger than 0.6 as positive samples and the rest as negative samples.

\subsubsection{Loss functions.}

The loss function of the SOT sub-net is as the following:
\begin{equation}
\begin{split}
    L_{sot} &= L_{sot}^{cls} + \lambda_{sot} L_{sot}^{reg},
\end{split}
    \label{eq:sotloss}
\end{equation}
where $\lambda_{sot}$ is a hyper-parameter for balancing the two loss functions. $L_{sot}^{cls}$ is the cross-entropy classification loss and $L_{sot}^{reg}$ is the regression loss.
%described below:
%\begin{equation}
%\begin{split}
%    L_{sot}^{reg} &= \sum_{i\in \{x,y,w,h\}}smooth_{L1}(\Delta[i])
%\end{split}
 %   \label{eq:sotlreg}
%\end{equation}
%where $\Delta$ is the predicted displacement for position and shape. $smooth_{L1}$ loss is the smoothed $L1$ loss defined in \cite{}.
% \begin{equation}
% \begin{split}
%     smooth_{L_1}(x,\sigma) = \left\{
%              \begin{array}{rcl}
%              \frac{1}{2}\sigma^2x^2, &  & {|x|< \frac{1}{\sigma^2}}\\
%              |x|-\frac{1}{2\sigma^2} &  & {|x|\ge \frac{1}{\sigma^2}}
%              \end{array}
%     \right.
% \end{split}
%     \label{eq:sotsml1}
% \end{equation}

We consider the re-identification task as a multi-class classification problem. A linear classifier is added after the backbone network and then the probability of each class is calculated through softmax. Finally we optimize the cross-entropy loss of the task:
\begin{equation}
\begin{split}
    L_{reid}(x, y) = -\sum_{i=0}^{N-1}y[i]\log(x[i])
\end{split}
    \label{eq:reidloss}
\end{equation}
where $N$ is the number of classes, $x$ and $y$ are network output and ground-truth, respectively. 
%\textcolor{red}{Some details on how the ground-truth is obtained, help others to understand how ReID net is learned.}

For the Newton boosting tree classifier and the quality filter, we use the loss function $L_{cls}(x, y) = -(1-y)\log(1-x)-y\log(x)$.

\subsection{Pre-Processing and Offline Clustering}
\begin{bf}
Detection score filter strategy.
\end{bf}
Sometimes the detection results given by the detector are extremely noisy, with strange false positives and the confidences are unreliable. We propose two solutions to the refine these detection results. The first one is a stricter NMS method. The second one is to train a confidence score refiner. Here we use the quality filter in long term cues selection as the confidence refiner. 

\begin{bf}
Long tracklets clustering
\end{bf}
Based on the output result of online mode, we design a simple batch clustering post-process procedure. For each tracklet, first we consider each frame as an isolated node, then if there are two nodes with similar ReID features, i.e., the cosine distance between them is less than a threshold, we will add an edge between the two nodes. Finally we can obtain some slices, each slice is one connected sub-graph. The second step is to merge these slices among different targets. Once again we merge two slices if they have small overlap in frame indexes, small spatial distance, and similar ReID features (we calculate mean feature distance of two slices). If two slices are merged, then the slices in the original place become new identities. Furthermore, after split and merge operations, we do interpolation in every tracklet in order to repair more false negatives.

%Thanks to the effective online MOT framework, this hand-crafted post-process is very fast because there are just a few slices.
\begin{table*}
\footnotesize
\begin{center}
\begin{tabular}{|l|c|c|c|c|c|c|c|c|c|}
\hline
Benchmark & Method & \textbf{MOTA} $\uparrow$ & MOTP $\uparrow$ & \textbf{IDF1} $\uparrow$ & IDP $\uparrow$ & IDR $\uparrow$ & FP $\downarrow$ & FN $\downarrow$ & IDS $\downarrow$ \\
\hline\hline
%MOT16 &  CDA\_DDALv2 & 43.9\% & 74.7\% & 45.1\% & 66.5\% & 34.1\% & 6450 & 95175 & 676 \\
%MOT16 &  TBSS & 44.6\% & 75.2\% & 42.6\% & 64.4\% & 31.9\% & 4136 & 96128 & 790 \\
%MOT16 & DCCRF16\cite{zhou2018deep} & 44.8\% & 75.6\% & 39.7\% & 58.4\% & 30.0\% & 5613 & 94133 & 968 \\
MOT16 & RAR16pub\cite{fang2018recurrent} & 45.9\% & 74.8\% & 48.8\% & 69.7\% & 37.5\% & 6871 & 91173 & 648 \\
MOT16 & STAM16\cite{chu2017online} & 46.0\% & 74.9\% & 50.0\% & 71.5\% & 38.5\% & 6895 & 91117 & 473 \\
MOT16 & DMMOT\cite{zhu2018online} & 46.1\% & 73.8\% & 54.8\% & 77.2\% & 42.5\% & 7909 & 89874 & 532 \\
MOT16 & AMIR\cite{sadeghian2017tracking} & 47.2\% & \textcolor[rgb]{1.00,0.00,0.00}{\textbf{75.8\%}} & 46.3\% & 68.9\% & 34.8\% & \textcolor[rgb]{1.00,0.00,0.00}{\textbf{2681}} & 92856 & 774 \\
MOT16 & MOTDT\cite{long2018real} & 47.6\% & 74.8\% & 50.9\% & 69.2\% & 40.3\% & 9253 & 85431 & 792 \\

MOT16 & Ours & 44.8\% & 75.1\% & 53.8\% & 75.2\% & 41.8\% & 9639 & 90571 & \textcolor[rgb]{1.00,0.00,0.00}{\textbf{451}} \\
MOT16 & Ours(with filter) & \textcolor[rgb]{1.00,0.00,0.00}{\textbf{49.2\%}} & 74.0\% & \textcolor[rgb]{1.00,0.00,0.00}{\textbf{56.5\%}} & \textcolor[rgb]{1.00,0.00,0.00}{\textbf{77.5\%}} & \textcolor[rgb]{1.00,0.00,0.00}{\textbf{44.5\%}} & 7187 & \textcolor[rgb]{1.00,0.00,0.00}{\textbf{84875}} & 606 \\
\hline\hline
MOT17 & PHD\_GSDL17\cite{fu2018particle} & 48.0\% & \textcolor[rgb]{1.00,0.00,0.00}{\textbf{77.2\%}} & 49.6\% & 68.4\% & 39.0\% & 23199 & 265954 & 3998 \\
MOT17 & AM\_ADM17\cite{lee2018learning} & 48.1\% & 76.7\% & 52.1\% & 71.4\% & 41.0\% & 25061 & 265495 & 2214 \\
MOT17 & DMAN\cite{zhu2018online} & 48.2\% & 75.7\% & 55.7\% & 75.9\% & 44.0\% & 26218 & 263608 & 2194 \\

MOT17 & HAM\_SADF17\cite{yoon2018online} & 48.3\% & \textcolor[rgb]{1.00,0.00,0.00}{\textbf{77.2\%}} & 51.1\% & 71.2\% & 39.9\% & \textcolor[rgb]{1.00,0.00,0.00}{\textbf{20967}} & 269038 & 1871 \\
MOT17 & MOTDT17\cite{long2018real} & 50.9\% & 76.6\% & 52.7\% & 70.4\% & 42.1\% & 24069 & 250768 & 2474 \\

MOT17 & Ours & 50.3\% & 76.8\% & 56.3\% & \textcolor[rgb]{1.00,0.00,0.00}{\textbf{76.5\%}} & 44.6\% & 21345 & 257062 & \textcolor[rgb]{1.00,0.00,0.00}{\textbf{1815}} \\
MOT17 & Ours(with filter) & \textcolor[rgb]{1.00,0.00,0.00}{\textbf{52.7\%}} & 76.2\% & \textcolor[rgb]{1.00,0.00,0.00}{\textbf{57.9\%}} & 76.3\% & \textcolor[rgb]{1.00,0.00,0.00}{\textbf{46.6\%}} & 22512 & \textcolor[rgb]{1.00,0.00,0.00}{\textbf{241936}} & 2167 \\
\hline\hline
MOT16p & EAMTT\_16\cite{sanchez2016online} & 52.5\% & 78.8\% & 53.3\% & 72.7\% & 42.1\% & \textcolor[rgb]{1.00,0.00,0.00}{\textbf{4407}} & 81223 & 910 \\
MOT16p & SORTwHPD16\cite{bewley2016simple} & 59.8\% & \textcolor[rgb]{1.00,0.00,0.00}{\textbf{79.6\%}} & 53.8\% & 65.2\% & 45.7\% & 8698 & 63245 & 1423 \\
MOT16p & DeepSORT\_2\cite{wojke2017simple} & 61.4\% & 79.1\% & 62.2\% & 72.1\% & 54.7\% & 12852 & 56668 & 781 \\
MOT16p & RAR16wVGG\cite{fang2018recurrent} & 63.0\% & 78.8\% & 63.8\% & 72.6\% & 56.9\% & 13663 & 53248 & \textcolor[rgb]{1.00,0.00,0.00}{\textbf{482}} \\
%MOT16p & TAP & 64.8\% & 78.6\% & \textcolor[rgb]{1.00,0.00,0.00}{\textbf{73.5\%}} & \textcolor[rgb]{1.00,0.00,0.00}{\textbf{82.7\%}} & \textcolor[rgb]{1.00,0.00,0.00}{\textbf{66.1\%}} & 13470 & 49927 & 794 \\
MOT16p & CNNMTT\cite{mahmoudi2018multi} & 65.2\% & 78.4\% & 62.2\% & 73.7\% & 53.8\% & 6578 & 55896 & 946 \\
MOT16p & POI\cite{yu2016poi} & 66.1\% & 79.5\% & 65.1\% & \textcolor[rgb]{1.00,0.00,0.00}{\textbf{77.7\%}} & 56.0\% & 5061 & 55914 & 805 \\

MOT16p & Ours & \textcolor[rgb]{1.00,0.00,0.00}{\textbf{69.6\%}} & 78.5\% & \textcolor[rgb]{1.00,0.00,0.00}{\textbf{68.6\%}} & 77.1\% & \textcolor[rgb]{1.00,0.00,0.00}{\textbf{61.7\%}} & 9138 & \textcolor[rgb]{1.00,0.00,0.00}{\textbf{45497}} & 768 \\
\hline
\end{tabular}
\end{center}
\caption{Comparision between the proposed MOT framework (online mode) with other online processing SOTA methods in MOT16 and MOT17. 'with filter' means detection score refiner is used. 'MOT16p' means MOT16 with private detection. \textcolor{red}{Red} for the best result.}
\label{tab:resol}
\end{table*}

\begin{table*}
\footnotesize
\begin{center}
\begin{tabular}{|l|c|c|c|c|c|c|c|c|c|c|}
\hline
Benchmark & Method & \textbf{MOTA} $\uparrow$ & MOTP $\uparrow$ & \textbf{IDF1} $\uparrow$ & IDP $\uparrow$ & IDR $\uparrow$ & FP $\downarrow$ & FN $\downarrow$ & IDS $\downarrow$ \\
\hline\hline
MOT17 & IOU17\cite{bochinski2017high} & 45.5\% & 76.9\% & 39.4\% & 56.4\% & 30.3\% & \textcolor[rgb]{1.00,0.00,0.00}{\textbf{19993}} & 281643 & 5988 \\
MOT17 & MHT\_bLSTM\cite{kim2018multi} & 47.5\% & \textcolor[rgb]{1.00,0.00,0.00}{\textbf{77.5\%}} & 51.9\% & 71.4\% & 40.8\% & 25981 & 268042 & 2069 \\
MOT17 & EDMT17\cite{chen2017enhancing} & 50.0\% & 77.3\% & 51.3\% & 67.0\% & 41.5\% & 32279 & 247297 & 2264 \\
MOT17 & MHT\_DAM\_17\cite{kim2015multiple} & 50.7\% & \textcolor[rgb]{1.00,0.00,0.00}{\textbf{77.5\%}} & 47.2\% & 63.4\% & 37.6\% & 22875 & 252889 & 2314 \\
MOT17 & jCC\cite{keuper2018motion} & 51.2\% & 75.9\% & 54.5\% & 72.2\% & 43.8\% & 25937 & 247822 & 1802 \\
MOT17 & FWT\_17\cite{henschel2018fusion} & 51.3\% & 77.0\% & 47.6\% & 63.2\% & 38.1\% & 24101 & 247921 & 2648 \\
MOT17 & Ours(with filter) & \textcolor[rgb]{1.00,0.00,0.00}{\textbf{54.7\%}} & 75.9\% & \textcolor[rgb]{1.00,0.00,0.00}{\textbf{62.3\%}} & \textcolor[rgb]{1.00,0.00,0.00}{\textbf{79.7\%}} & \textcolor[rgb]{1.00,0.00,0.00}{\textbf{51.1\%}} & 26091 & \textcolor[rgb]{1.00,0.00,0.00}{\textbf{228434}} & \textcolor[rgb]{1.00,0.00,0.00}{\textbf{1243}} \\
\hline\hline
%MOT16p & IOU & 57.1\% & 77.1\% & 46.9\% & 59.8\% & 38.6\% & 5702 & 70278 & 2167 \\
MOT16p & NOMTwSDP16\cite{choi2015near} & 62.2\% & 79.6\% & 62.6\% & 77.2\% & 52.6\% & \textcolor[rgb]{1.00,0.00,0.00}{\textbf{5119}} & 63352 & 406 \\
MOT16p & MCMOT\_HDM\cite{lee2016multi} & 62.4\% & 78.3\% & 51.6\% & 60.7\% & 44.9\% & 9855 & 57257 & 1394 \\
MOT16p & KDNT\cite{yu2016poi} & 68.2\% & 79.4\% & 60.0\% & 66.9\% & 54.4\% & 11479 & 45605 & 933 \\
% MOT16p & LM\_NN & 69.0\% & 79.9\% & 61.9\% & 71.6\% & 54.5\% & 6213 & 49675 & 668 \\
MOT16p & LMP\_p\cite{tang2017multiple} & 71.0\% & \textcolor[rgb]{1.00,0.00,0.00}{\textbf{80.2\%}} & 70.1\% & 78.9\% & 63.0\% & 7880 & 44564 & \textcolor[rgb]{1.00,0.00,0.00}{\textbf{434}} \\
MOT16p & HT\_SJTUZTE\cite{lin2017real} & \textcolor[rgb]{1.00,0.00,0.00}{\textbf{71.3\%}} & 79.3\% & 67.6\% & 75.2\% & 61.4\% & 9238 & 42521 & 617 \\

MOT16p & Ours & 71.2\% & 78.3\% & \textcolor[rgb]{1.00,0.00,0.00}{\textbf{73.1\%}} & \textcolor[rgb]{1.00,0.00,0.00}{\textbf{80.7\%}} & \textcolor[rgb]{1.00,0.00,0.00}{\textbf{66.8\%}} & 10274 & \textcolor[rgb]{1.00,0.00,0.00}{\textbf{41732}} & 510 \\
\hline
\end{tabular}
\end{center}
\caption{Comparision between the proposed MOT framework (batch mode) with other batch processing SOTA methods in MOT16 and MOT17. 'with filter' means detection score refiner is used. 'MOT16p' means MOT16 with private detection. \textcolor{red}{Red} for the best result.}
\label{tab:resof}
\end{table*}

%------------------------------------------------------------------------
\section{Experiments}
%-------------------------------------------------------------------------
\subsection{Implement Details}
This framework is written in Python with PyTorch support. All CNNs are pre-trained on Imagenet dataset and then trained on MOT task. Additionally, we use extra private data in training two sub-net but only public data in training the classifier and the quality filter. The public data we use is 7 video sequences from MOT16 benchmark data. The private data contain labeled videos of pedestrians which are quite different from all testset videos. 
The amount of the private data for trainning is about 100 minutes of 25 fps videos and the number of different pedestrians is about 1000.

\begin{bf}
Tracklet status update.
\end{bf}
For matched tracklets, their positions will be updated as the position of the matched detection result and the SOT template will be updated. Otherwise, the position is kept unchanged, and so do the template.
Despite the matching results, Kalman Filter is used to smooth the trajectories. The quality decay value is 0.95 and the exponent $k$ = 16. When the overall quality score is lower than 0.5, the target will not be output. The matching threshold $\zeta_m = 0.05$ and drop threshold $\zeta_t = 0.1$.

\begin{bf}
Training.
\end{bf}
We use similar approach as the original work in the SOT sub-net but change anchors to $[1.0, 1.6, 2.2, 2.8, 3.4]$.  When training the ReID sub-net, we use the Stochastic Gradient Descent (SGD) optimizer and set the initial learning rate to 0.1, decay by 0.5 every 8 epoches for 96 epoches in total. The ReID sub-net is trained with weight decay of $5\times10^{-4}$ and mini-batch size of $32$.

We train the swithcer aware classifier under xgboost framework\cite{chen2016xgboost}. The number of trees is set to 410. The max depth is 5, learning rate is 0.05 and the minimum leaf node weight is set to 1.

For long term cues, the selection interval of the tracklet history can be set from 10 to 20. We just simply sample $K=3$ frames and set $\delta = 15$.
\subsection{Evaluation on MOT Benchmarks}

\begin{bf}
Datasets.
\end{bf}
The proposed framework is evaluated under the MOT16 and MOT17 benchmarks \cite{MOT16}. They share the same test videos but offer different detection input. MOT17 has fixed the ground truth and make them more accurate. The test video sequences include various complicated scenes and are still a great challenge.

\begin{bf}
Evaluate Metrics.
\end{bf}
Following the benchmarks, we evaluate our work with the CLEAR MOT Metrics\cite{bernardin2008evaluating}. Among the metrics, MOTA and IDF1 are regarded the most important. MOTA conclude\ the recall, precision as well as id-switch count. IDF1 indicates the average maximum consistent tracking rate.
In evaluation of MOT trackers, the criteria MOTA is deeply relevant to the detector recall and precision, while ID-Switch impresses far less than them. However IDF1 can indicate the consistency. A robust tracking system should have both high MOTA and IDF1 score.

\begin{bf}
Results.
\end{bf}
Table~\ref{tab:resol} and Table~\ref{tab:resof} show the results of online and batch processing methods both in MOT16 and MOT17, respectively. Besides, in task of MOT16 with private detector, our tracker, as well as KDNT, LMP\_p, HT\_SJTUZTE and POI trackers use the same detector that proposed by the author of POI tracker. The detections are available on google drive\footnote[1]{\scriptsize https://drive.google.com/open?id=0B5ACiy41McAHMjczS2p0dFg3emM}.

The results show that our framework outperforms many previous the state-of-the-art trackers in both MOT16 and MOT17 benchmarks. Both MOTA and IDF1 scores are in the leading position in MOT16/MOT17 among online/batch processing algorithms. Our batch processing algorithm achieved the highest MOTA in MOT17 benchmark\footnote[1]{Up to the date of 14/11/2018.}.
\subsection{Ablation Study and Discussion}
\label{sec:abla}
\begin{figure}
\begin{center}
%\fbox{\rule{0pt}{2in} \rule{.9\linewidth}{0pt}}
\includegraphics[width=1.0\linewidth]{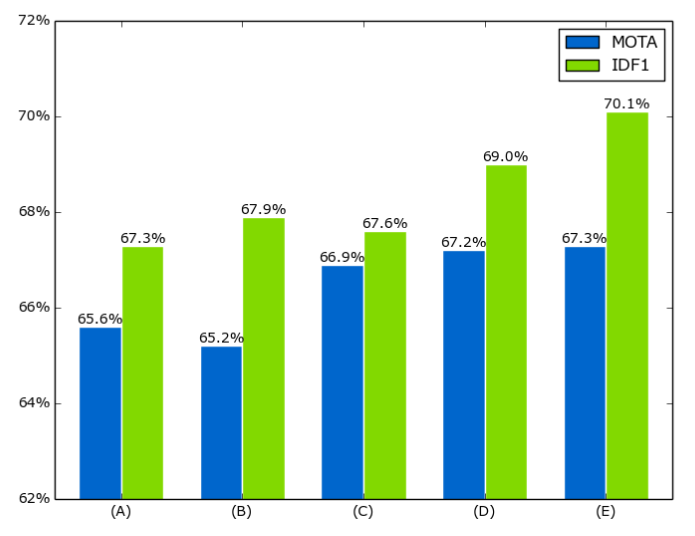}
\end{center}
   \caption{Analysis of our framework using different components. (A) baseline (hand-crafted) (B) long-term cues only (C) short-term cues only (D) long and short term cues (E) long and short term cues with SAC}
\label{fig:cues}
\end{figure}
\begin{bf}
How do different cues influence the tracking quality?
\end{bf}

The ablation study was evaluated on MOT16 trainset. Since we have used the trainset for validation, we exclude the MOT16 trainset from the training data when training the sub-nets, the switcher-aware classifier and the quality filter for all ablation study results in Figure~\ref{fig:cues}. For the classifier and quality filter, we use extra private training data in ablation study. 

Figure~\ref{fig:cues} shows the impact of different components. The baseline model, (A) in Fig. \ref{fig:cues}, does not use a learned classifier but calculates affinity in a hand-crafted way using only position information. The other experimental results in Fig. \ref{fig:cues} share the same settings of framework except the input features for classifier and four different classifiers are trained to fit the different input features. It can be seen from the figure that the short term cues provide more improvement for MOTA than long term cues when compared with the baseline. It is also the most intuitive reflection of the relevance between the two adjacent detection frames. When short term cues are utilized effectively, the MOTA score is improved by 1.3\% and IDF1 is improved by 0.3 \%. 
%In most cases we can make correct matching through short term cues.
On the other hand, combining long term cues can effectively improve the discriminative ability between the tracklets, which brings increment of IDF1 by 1.4\%. However, the MOTA improvement from long term cues is less than the improvement from the short term cues. Combining both short term and long term cues performs better than using single cue, which validates that these two cues are complementary to each other.
Thirdly, adding the switcher-aware classifier can greatly reduce id-switches number, which leads to another 1.1\% increment in IDF1, while it has just a little effect on MOTA.
The learning approach to combine long term and short term cues is effective and the combination using SAC brings improvements on multiple metrics.

\begin{bf}
How does SAC work?
\end{bf}

\begin{figure}
\begin{center}
%\fbox{\rule{0pt}{2in} \rule{.9\linewidth}{0pt}}
\includegraphics[width=1.0\linewidth]{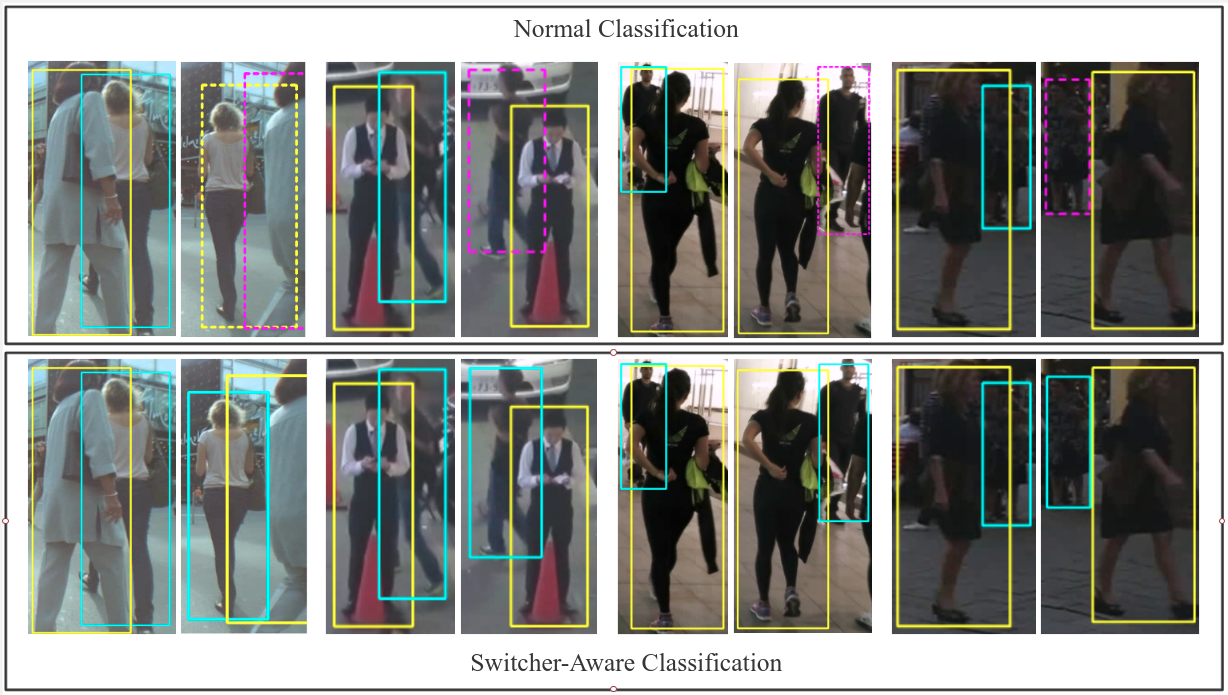}
\end{center}
   \caption{Examples of identity switch cases (top row) fixed when SAC is used (bottom row) for matching targets between adjacent frames. Dashed line boxes indicate id-switch. }
\label{fig:fixcase}
\end{figure}

We also analyze the real effects on the videos. Figure~\ref{fig:fixcase} shows that with switcher-aware classification, the tracking is more robust. The main contribution of SAC is that it fixes a lot of id-switches. After SAC is used, in MOT16 private the id-switch number decreases from 642 to 569, IDF1 increases by 1.1\%. This is because in traditional pairwise matching, lack of comparison to local switcher brings mistakes when the pair is occluded and therefore judged not to match. Besides, when occlusion happens, SAC helps to discriminate different targets.

\begin{bf}
Can multiple cues been handle in one network?
\end{bf}

We have tried to extract some of the features from SOT backbone CNN and combine with the ReID branch through a ROI-Pooling layer. The experiment shows that a multi-task training of ReID and SOT task leads to drop in both SOT and ReID accuracy. Replacing Siamese-RPN by the multi-task trained network in MOT task, the MOTA decreases by 0.6\% and IDF1 decreases by 2.5\%. The SOT task needs background knowledge while the ReID task wants to eliminate it, which causes conflict during feature learning if a single network is used for both tasks. For the time being it is hard to handle two cues in one single network, further research works need to be done.

\begin{bf}
Why do we need small feature dimension?
\end{bf}

We use a input feature of small dimension to balance the feature length of different information. It is well-known that motion and position features are usually very short. If we combine them with long appearance features, it is hard to fully utilize the position and motion features. We have tried directly concatenating raw ReID feature with short position features, the experimental result shows that imbalanced input features decreases IDF1 by 1.3\%, MOTA by 0.1\%  and the IDS number increases by 88.
%In previous work like \cite{sadeghian2017tracking}, they concatenate the motion features with raw appearance features, this makes the input of the network imbalanced. 
We believe that the position and motion information are important and they should be emphasized by reducing the dimension of appearance features.
Another inevitable issue is the complexity of data association. Short features make the procedure faster.
%It seems that the procedure is $O(n)$ theoretically, but actually a constant factor $C$, which denotes the average match times for single target, must be multiplied, and usually $C$ can be large in crowded scenes. 

\begin{bf}
Why we use boosting trees for classification?
\end{bf}

We have tried other different classifiers like neural network of linear layers(NN), support vector machine (SVM), but the boosting decision tree (BDT) is the best one. This experiment was done using full training data include MOT16. The MOTA of NN, SVM and BDT are 67.0\%, 67.6\% and 67.8\%, respectively. We have found that for such a small input feature, neural network does not perform well on such small scale dataset.
% In the data association step of our framework, short but meaningful features are used and the complexity of each matching calculation will cost less time than CNNs or RNNs.

% \begin{bf}
% What is the difference if another SOT sub-net is used?
% \end{bf}

%------------------------------------------------------------------------
\section{Conclusions}
In this paper, we have presented an effective MOT framework that learns to combine long term and short term cues. The long term cues can help correct the mistakes of short term cues, e.g., avoid the SOT sub-net to drift during occlusion. We also propose a switcher-aware classification method to improve the robustness of tracking system. The outstanding performance on MOT benchmarks demonstrates the effectiveness of the proposed framework.

{\small
\bibliographystyle{ieee}
\bibliography{egbib}

\begin{thebibliography}{10}\itemsep=-1pt

\bibitem{bae2014robust}
S.-H. Bae and K.-J. Yoon.
\newblock Robust online multi-object tracking based on tracklet confidence and
  online discriminative appearance learning.
\newblock In {\em Proceedings of the IEEE conference on computer vision and
  pattern recognition}, pages 1218--1225, 2014.

\bibitem{bae2018confidence}
S.-H. Bae and K.-J. Yoon.
\newblock Confidence-based data association and discriminative deep appearance
  learning for robust online multi-object tracking.
\newblock {\em IEEE transactions on pattern analysis and machine intelligence},
  40(3):595--610, 2018.

\bibitem{bernardin2008evaluating}
K.~Bernardin and R.~Stiefelhagen.
\newblock Evaluating multiple object tracking performance: the clear mot
  metrics.
\newblock {\em Journal on Image and Video Processing}, 2008:1, 2008.

\bibitem{bertinetto2016fully}
L.~Bertinetto, J.~Valmadre, J.~F. Henriques, A.~Vedaldi, and P.~H. Torr.
\newblock Fully-convolutional siamese networks for object tracking.
\newblock In {\em European conference on computer vision}, pages 850--865.
  Springer, 2016.

\bibitem{bewley2016simple}
A.~Bewley, Z.~Ge, L.~Ott, F.~Ramos, and B.~Upcroft.
\newblock Simple online and realtime tracking.
\newblock In {\em Image Processing (ICIP), 2016 IEEE International Conference
  on}, pages 3464--3468. IEEE, 2016.

\bibitem{bochinski2017high}
E.~Bochinski, V.~Eiselein, and T.~Sikora.
\newblock High-speed tracking-by-detection without using image information.
\newblock In {\em Advanced Video and Signal Based Surveillance (AVSS), 2017
  14th IEEE International Conference on}, pages 1--6. IEEE, 2017.

\bibitem{chen2017enhancing}
J.~Chen, H.~Sheng, Y.~Zhang, and Z.~Xiong.
\newblock Enhancing detection model for multiple hypothesis tracking.
\newblock In {\em Conf. on Computer Vision and Pattern Recognition Workshops},
  pages 2143--2152, 2017.

\bibitem{chen2016xgboost}
T.~Chen and C.~Guestrin.
\newblock Xgboost: A scalable tree boosting system.
\newblock In {\em Proceedings of the 22nd acm sigkdd international conference
  on knowledge discovery and data mining}, pages 785--794. ACM, 2016.

\bibitem{choi2015near}
W.~Choi.
\newblock Near-online multi-target tracking with aggregated local flow
  descriptor.
\newblock In {\em Proceedings of the IEEE international conference on computer
  vision}, pages 3029--3037, 2015.

\bibitem{chu2017online}
Q.~Chu, W.~Ouyang, H.~Li, X.~Wang, B.~Liu, and N.~Yu.
\newblock Online multi-object tracking using cnn-based single object tracker
  with spatial-temporal attention mechanism.
\newblock In {\em 2017 IEEE International Conference on Computer Vision
  (ICCV).(Oct 2017)}, pages 4846--4855, 2017.

\bibitem{danelljan2017eco}
M.~Danelljan, G.~Bhat, F.~S. Khan, M.~Felsberg, et~al.
\newblock Eco: Efficient convolution operators for tracking.
\newblock In {\em CVPR}, page~3, 2017.

\bibitem{fang2018recurrent}
K.~Fang, Y.~Xiang, X.~Li, and S.~Savarese.
\newblock Recurrent autoregressive networks for online multi-object tracking.
\newblock In {\em 2018 IEEE Winter Conference on Applications of Computer
  Vision (WACV)}, pages 466--475. IEEE, 2018.

\bibitem{fu2018particle}
Z.~Fu, P.~Feng, F.~Angelini, J.~Chambers, and S.~M. Naqvi.
\newblock Particle phd filter based multiple human tracking using online
  group-structured dictionary learning.
\newblock {\em IEEE Access}, 6:14764--14778, 2018.

\bibitem{held2016learning}
D.~Held, S.~Thrun, and S.~Savarese.
\newblock Learning to track at 100 fps with deep regression networks.
\newblock In {\em European Conference Computer Vision (ECCV)}, 2016.

\bibitem{henschel2018fusion}
R.~Henschel, L.~Leal-Taix{\'e}, D.~Cremers, and B.~Rosenhahn.
\newblock Fusion of head and full-body detectors for multi-object tracking.
\newblock In {\em Computer Vision and Pattern Recognition Workshops (CVPRW)},
  2018.

\bibitem{huang2008robust}
C.~Huang, B.~Wu, and R.~Nevatia.
\newblock Robust object tracking by hierarchical association of detection
  responses.
\newblock In {\em European Conference on Computer Vision}, pages 788--801.
  Springer, 2008.

\bibitem{keuper2018motion}
M.~Keuper, S.~Tang, B.~Andres, T.~Brox, and B.~Schiele.
\newblock Motion segmentation \& multiple object tracking by correlation
  co-clustering.
\newblock {\em IEEE transactions on pattern analysis and machine intelligence},
  2018.

\bibitem{kim2015multiple}
C.~Kim, F.~Li, A.~Ciptadi, and J.~M. Rehg.
\newblock Multiple hypothesis tracking revisited.
\newblock In {\em Proceedings of the IEEE International Conference on Computer
  Vision}, pages 4696--4704, 2015.

\bibitem{kim2018multi}
C.~Kim, F.~Li, and J.~M. Rehg.
\newblock Multi-object tracking with neural gating using bilinear lstm.
\newblock In {\em Proceedings of the European Conference on Computer Vision
  (ECCV)}, pages 200--215, 2018.

\bibitem{lee2016multi}
B.~Lee, E.~Erdenee, S.~Jin, M.~Y. Nam, Y.~G. Jung, and P.~K. Rhee.
\newblock Multi-class multi-object tracking using changing point detection.
\newblock In {\em European Conference on Computer Vision}, pages 68--83.
  Springer, 2016.

\bibitem{lee2018learning}
S.-H. Lee, M.-Y. Kim, and S.-H. Bae.
\newblock Learning discriminative appearance models for online multi-object
  tracking with appearance discriminability measures.
\newblock {\em IEEE Access}, Early Access:1--1, 2018.

\bibitem{li2018high}
B.~Li, J.~Yan, W.~Wu, Z.~Zhu, and X.~Hu.
\newblock High performance visual tracking with siamese region proposal
  network.
\newblock In {\em Proceedings of the IEEE Conference on Computer Vision and
  Pattern Recognition}, pages 8971--8980, 2018.

\bibitem{lin2017real}
W.~Lin, J.~Peng, S.~Deng, M.~Liu, X.~Jia, and H.~Xiong.
\newblock Real-time multi-object tracking with hyper-plane matching.
\newblock Technical report, Shanghai Jiao Tong University and ZTE Corp, 2017.

\bibitem{long2018real}
C.~Long, A.~Haizhou, Z.~Zijie, and S.~Chong.
\newblock Real-time multiple people tracking with deeply learned candidate
  selection and person re-identification.
\newblock ICME, 2018.

\bibitem{mahmoudi2018multi}
N.~Mahmoudi, S.~M. Ahadi, and M.~Rahmati.
\newblock Multi-target tracking using cnn-based features: Cnnmtt.
\newblock {\em Multimedia Tools and Applications}, pages 1--20, 2018.

\bibitem{MOT16}
A.~Milan, L.~Leal-Taix\'{e}, I.~Reid, S.~Roth, and K.~Schindler.
\newblock {MOT}16: {A} benchmark for multi-object tracking.
\newblock {\em arXiv:1603.00831 [cs]}, Mar. 2016.
\newblock arXiv: 1603.00831.

\bibitem{milan2017online}
A.~Milan, S.~H. Rezatofighi, A.~R. Dick, I.~D. Reid, and K.~Schindler.
\newblock Online multi-target tracking using recurrent neural networks.
\newblock In {\em AAAI}, volume~2, page~4, 2017.

\bibitem{Munkres1957Algorithms}
J.~Munkres.
\newblock Algorithms for the assignment and transportation problems.
\newblock {\em Journal of the Society for Industrial \& Applied Mathematics},
  5(1):32--38, 1957.

\bibitem{pirsiavash2011globally}
H.~Pirsiavash, D.~Ramanan, and C.~C. Fowlkes.
\newblock Globally-optimal greedy algorithms for tracking a variable number of
  objects.
\newblock In {\em Computer Vision and Pattern Recognition (CVPR), 2011 IEEE
  Conference on}, pages 1201--1208. IEEE, 2011.

\bibitem{sadeghian2017tracking}
A.~Sadeghian, A.~Alahi, and S.~Savarese.
\newblock Tracking the untrackable: Learning to track multiple cues with
  long-term dependencies.
\newblock In {\em 2017 IEEE International Conference on Computer Vision
  (ICCV)}, pages 300--311. IEEE, 2017.

\bibitem{sanchez2016online}
R.~Sanchez-Matilla, F.~Poiesi, and A.~Cavallaro.
\newblock Online multi-target tracking with strong and weak detections.
\newblock In {\em European Conference on Computer Vision}, pages 84--99.
  Springer, 2016.

\bibitem{son2017multi}
J.~Son, M.~Baek, M.~Cho, and B.~Han.
\newblock Multi-object tracking with quadruplet convolutional neural networks.
\newblock In {\em Proceedings of the IEEE Conference on Computer Vision and
  Pattern Recognition}, pages 5620--5629, 2017.

\bibitem{tang2015subgraph}
S.~Tang, B.~Andres, M.~Andriluka, and B.~Schiele.
\newblock Subgraph decomposition for multi-target tracking.
\newblock In {\em Proceedings of the IEEE Conference on Computer Vision and
  Pattern Recognition}, pages 5033--5041, 2015.

\bibitem{tang2016multi}
S.~Tang, B.~Andres, M.~Andriluka, and B.~Schiele.
\newblock Multi-person tracking by multicut and deep matching.
\newblock In {\em European Conference on Computer Vision}, pages 100--111.
  Springer, 2016.

\bibitem{tang2017multiple}
S.~Tang, M.~Andriluka, B.~Andres, and B.~Schiele.
\newblock Multiple people tracking by lifted multicut and person
  reidentification.
\newblock In {\em Proceedings of the IEEE Conference on Computer Vision and
  Pattern Recognition}, pages 3539--3548, 2017.

\bibitem{wojke2017simple}
N.~Wojke, A.~Bewley, and D.~Paulus.
\newblock Simple online and realtime tracking with a deep association metric.
\newblock In {\em Image Processing (ICIP), 2017 IEEE International Conference
  on}, pages 3645--3649. IEEE, 2017.

\bibitem{xiang2015learning}
Y.~Xiang, A.~Alahi, and S.~Savarese.
\newblock Learning to track: Online multi-object tracking by decision making.
\newblock In {\em Proceedings of the IEEE international conference on computer
  vision}, pages 4705--4713, 2015.

\bibitem{yan2012track}
X.~Yan, X.~Wu, I.~A. Kakadiaris, and S.~K. Shah.
\newblock To track or to detect? an ensemble framework for optimal selection.
\newblock In {\em European Conference on Computer Vision}, pages 594--607.
  Springer, 2012.

\bibitem{yoon2018online}
Y.-c. Yoon, A.~Boragule, K.~Yoon, and M.~Jeon.
\newblock Online multi-object tracking with historical appearance matching and
  scene adaptive detection filtering.
\newblock {\em arXiv preprint arXiv:1805.10916}, 2018.

\bibitem{yu2016poi}
F.~Yu, W.~Li, Q.~Li, Y.~Liu, X.~Shi, and J.~Yan.
\newblock Poi: Multiple object tracking with high performance detection and
  appearance feature.
\newblock In {\em European Conference on Computer Vision}, pages 36--42.
  Springer, 2016.

\bibitem{zhang2008global}
L.~Zhang, Y.~Li, and R.~Nevatia.
\newblock Global data association for multi-object tracking using network
  flows.
\newblock In {\em Computer Vision and Pattern Recognition, 2008. CVPR 2008.
  IEEE Conference on}, pages 1--8. IEEE, 2008.

\bibitem{zhu2018online}
J.~Zhu, H.~Yang, N.~Liu, M.~Kim, W.~Zhang, and M.-H. Yang.
\newblock Online multi-object tracking with dual matching attention networks.
\newblock In {\em Proceedings of the European Conference on Computer Vision
  (ECCV)}, pages 366--382, 2018.

\end{thebibliography}
}

\end{document}